\def\year{2021}\relax
\documentclass[letterpaper]{article} % DO NOT CHANGE THIS
\usepackage{aaai21}  % DO NOT CHANGE THIS
\usepackage{times}  % DO NOT CHANGE THIS
\usepackage{helvet} % DO NOT CHANGE THIS
\usepackage{courier}  % DO NOT CHANGE THIS
\usepackage[hyphens]{url}  % DO NOT CHANGE THIS
\usepackage{graphicx} % DO NOT CHANGE THIS
\usepackage{hyperref}
\hypersetup{
    colorlinks=true,
    linkcolor=black,
    filecolor=black,      
    urlcolor=blue,
    citecolor=black,
}
\usepackage{natbib}  % DO NOT CHANGE THIS AND DO NOT ADD ANY OPTIONS TO IT
\usepackage{caption} % DO NOT CHANGE THIS AND DO NOT ADD ANY OPTIONS TO IT
\title{Temporal ROI Align for Video Object Recognition}
\author {
        Tao Gong \textsuperscript{\rm 1,}\textsuperscript{\rm 2}\thanks{Work done during an internship at SenseTime.},
        Kai Chen \textsuperscript{\rm 3},
        Xinjiang Wang \textsuperscript{\rm 3},
				 Qi Chu \textsuperscript{\rm 1,}\textsuperscript{\rm 2}\thanks{Qi Chu is the corresponding author.},
				 Feng Zhu \textsuperscript{\rm 3},\\
        Dahua Lin \textsuperscript{\rm 4},
        Nenghai Yu \textsuperscript{\rm 1,}\textsuperscript{\rm 2},
				 Huamin Feng \textsuperscript{\rm 5}\\
}
\affiliations {
    \textsuperscript{\rm 1} School of Cyberspace Security, University of Science and Technology of China\\
    \textsuperscript{\rm 2}Key Laboratory of Electromagnetic Space Information, Chinese Academy of Sciences\\
    \textsuperscript{\rm 3} SenseTime Research\\
    \textsuperscript{\rm 4} The Chinese University of Hong Kong\\
    \textsuperscript{\rm 5} Beijing Electronic Science and Technology Institute  \\
gt950513@mail.ustc.edu.cn, 
chenkai@sensetime.com, 
wangxinjiang@sensetime.com, 
qchu@ustc.edu.cn, 
zhufeng@sensetime.com, 
dhlin@ie.cuhk.edu.hk, 
ynh@ustc.edu.cn, 
oliver\_feng@yeah.net
}

\usepackage{amsmath}
\usepackage{amssymb}
\usepackage{amsthm}
\usepackage{makecell}
\usepackage[switch]{lineno}
\begin{document}
%\linenumbers

\maketitle

\begin{abstract}
%\vspace{-0.2cm}
Video object detection is challenging in the presence of appearance deterioration in certain video frames. Therefore, it is a natural choice to aggregate temporal information from other frames of the same video into the current frame. However, ROI Align, as one of the most core procedures of video detectors, still remains extracting features from a single-frame feature map for proposals, making the extracted ROI features lack temporal information from videos. In this work, considering the features of the same object instance are highly similar among frames in a video, a novel Temporal ROI Align operator is proposed to extract features from other frames feature maps for current frame proposals by utilizing feature similarity. The proposed Temporal ROI Align operator can extract temporal information from the entire video for proposals. We integrate it into single-frame video detectors and other state-of-the-art video detectors, and conduct quantitative experiments to demonstrate that the proposed Temporal ROI Align operator can consistently and significantly boost the performance. Besides, the proposed Temporal ROI Align can also be applied into video instance segmentation. Codes are available at \href{https://github.com/open-mmlab/mmtracking}{https://github.com/open-mmlab/mmtracking}.
\end{abstract}

\vspace{-0.35cm}
\section{Introduction}
%\vspace{-0.25cm}
Recently, deep convolutional neural networks have brought great progress in object detection of still images \cite{ren2015faster,lin2017focal,Cai_2018_CVPR,tian2019fcos,duan2019centernet}. Most state-of-the-art single image object detectors usually adopt the region-based detection paradigm \cite{ren2015faster,Cai_2018_CVPR}.
When directly applying these detectors for video object detection (VID), the detection accuracy suffers from deteriorated object appearances in videos, such as motion blur, video defocus and object occlusions. Despite these challenges, video contains temporal information about the same object instance (e.g., its appearance in different poses, and from different viewpoints). Therefore, the key challenge is how to effectively exploit such temporal information for the same object instance in a video.

\begin{figure}[t]
\centering
\includegraphics[width=8.4cm]{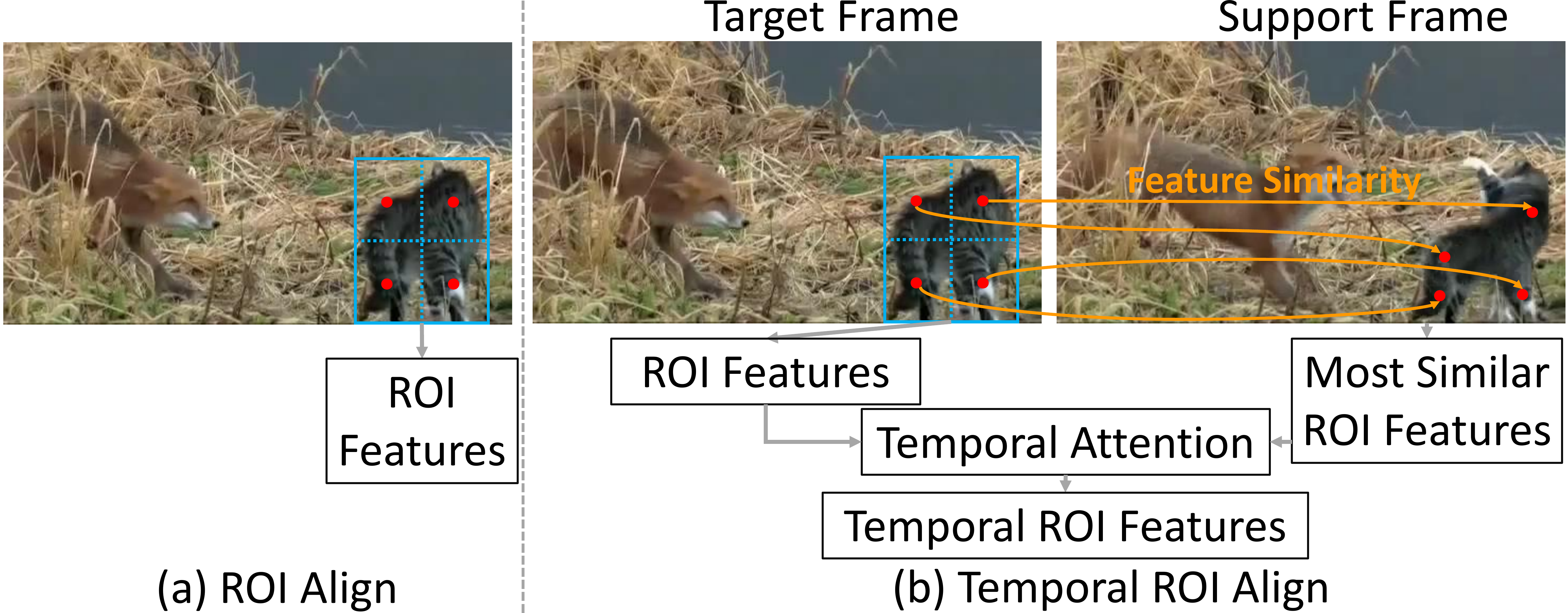}
\vspace{-0.25cm}
\caption{The Illustration of Temporal ROI Align. (a) The conventional ROI Align. The spatial size of ROI features is set to $2\times2$ for convenience. (b) Based on feature similarity, Temporal ROI Align implicitly extracts the most similar ROI features from support frames feature maps for target frame proposals. Then, temporal attention mechanism is conducted to aggregate the ROI features and the most similar ROI features in order to generate final temporal ROI features}
\label{fig:motivation}
\vspace{-0.35cm}
\end{figure}

As shown in Fig. \ref{fig:motivation} (a), region-based detectors usually adopt ROI Align \cite{He_2017_ICCV} to extract ROI features. However, ROI Align only utilizes current frame feature map to extract features for current frame proposals, this leads to the extracted ROI features lacking the temporal information of the same object instance in a video. One simple and obvious way to utilize the temporal information is taking other frames feature maps to perform ROI Align for current frame proposals. But the precise location of the current frame proposals in other frame feature maps is unknown, making the simple method infeasible.

In fact, there are already many previous works \cite{zhu2017flow,wu2019sequence,xiao2018video,liu2018mobile,chen2018optimizing,deng2019object} attempting to exploit temporal information in videos. Some works try to utilize image level information from nearby frames to help detection. For example, FGFA \cite{zhu2017flow}, MANet \cite{wang2018fully} utilize optical flow \cite{dosovitskiy2015flownet}, and STSN \cite{bertasius2018object} applies deformable convolutions \cite{Dai_2017_ICCV} for image feature calibration. PS ROI Pooling \cite{dai2016r} is used to pool features for proposals from the calibrated image feature map in these methods. The extracted PS ROI features contain temporal information from nearby frames. However, these methods can only utilize  nearby frames within one second (usually at most 30 frames). Performance will degrade with longer time interval, making it hard to leverage information from frames that are far apart in time. Other works attempt to utilize proposal level information from longer video length. SELSA \cite{wu2019sequence} and \cite{shvets2019leveraging} aggregate the high-level proposal features (fully connection layer features of proposals) with each other in order to make every proposal features in current frame contain the high-level proposal features from other frames. However, the ROI features are still extracted from a single image.

In this paper, a novel operator, named as Temporal ROI Align, is proposed to exploit the temporal information for the same object instance in a video. As shown in Fig. \ref{fig:motivation} (b), we define a target frame as a frame where final prediction is done at the moment. The target frame is allowed to have multiple support frames, which are used for strengthening the features of the target  frame. Considering the features of the same object instance are highly similar among frames in a video, the proposed operator implicitly extracts the most similar ROI features from support frames feature maps for target frame proposals based on feature similarity. The extracted most similar ROI features contain the temporal information of the same object instance in a video. The key challenge now is how to effectively aggregate these ROI features. One simple and obvious way is to average them. However, it is suboptimal due to the fact that an object instance may be blurry in some frames and clear in other frames. It is obvious that the ROI features of clear object instances should play a more important role than features of blurry object instances during aggregation. Therefore, temporal attention mechanism is conducted to aggregate the ROI features and the most similar ROI features.

The proposed operator also inherits the advantage of SELSA \cite{wu2019sequence} and \cite{shvets2019leveraging} which leverage long-range temporal information of a video. It can also be applied into other video object detectors and other video tasks of computer vision, such as video instance segmentation (VIS).

The contributions of the paper are summarized as follows:

1. A novel Temporal ROI Align operator is proposed to offer an alternative for the conventional ROI Align operator in videos. It can extract temporal information from an entire video with arbitrary length for a proposal.

2. We integrate the proposed Temporal ROI Align into single-frame video detectors and other state-of-the-art video detectors, and demonstrate through quantitative experiments that the proposed Temporal ROI Align operator can consistently and significantly boost the performance.

%2. Experiments show that the proposed Temporal ROI Align can not only be integrated into single-frame video detectors, but also be integrated into rencently proposed video object detectors to further improve the performance.

3. Experiments show that the proposed Temporal ROI Align can also be used in other video tasks of computer vision, such as video instance segmentation (VIS).
%\begin{enumerate}
%\item A novel Temporal ROI Align operator is proposed to offer an alternative for the conventional ROI Align operator in videos. It can extract temporal information for a proposal from a entire video with arbitrary length.
%\item Experiments show that the proposed Temporal ROI Align can not only be integrated into single-frame video detectors, but also be integrated into rencently proposed video object detectors to further improve the performance.
%\item Experiments show that the proposed Temporal ROI Align can also be used in other computer vision tasks, such as video instance segmentation (VIS).
%\end{enumerate}

%\vspace{-0.25cm}
\section{Related Work}
%\vspace{-0.25cm}
%This section briefly reviews several works that are closely related to this work.
\subsection{Object Detection In Still Images}
%With the development of the deep convolutional neural networks, the object detectors using deep convolutional neural networks \cite{girshick2014rich,girshick2015fast,ren2015faster,redmon2016you,liu2016ssd,dai2016r} improve the accuracy of the object detection dramatically.
%Although there are some one stage object detectors \cite{liu2016ssd,redmon2016you,Redmon_2017_CVPR,tian2019fcos,duan2019centernet} been proposed,
Most state-of-the-art object detectors \cite{girshick2014rich,girshick2015fast,ren2015faster,dai2016r} usually adopt two stage framework composed of a proposal generator and a region classifier. Fast R-CNN \cite{girshick2015fast} develops a ROI Pooling layer to extract features of proposals as the input of region classifier. Faster R-CNN \cite{ren2015faster} proposes Region Proposal Network (RPN) to combine proposal generation with region classification into one framework.
%R-FCN \cite{dai2016r} replaces ROI pooling with PS ROI pooling to extract features of proposal.
%Cascade R-CNN \cite{Cai_2018_CVPR} builds the multi-stage box classification and the multi-stage box regression to get better localization of the objects.
The methods mentioned above usually adopt ROI Align \cite{He_2017_ICCV} or ROI Pooling to extract features for proposals within a single image. As a consequence, when applying these single image detectors to VID, the features extracted from ROI Align lack the temporal information in videos.

Recently, self-attention mechanisms \cite{vaswani2017attention} have been proven to be beneficial for single image recognition. Object relation module \cite{hu2018relation} is proposed to model the proposal relationship within a single image. \cite{gu2018learning} proposes a method which can extract features for proposals using the features of a whole image rather than only the features covered by proposals. Different from these methods which attempt to interact high-level features among objects (e.g. computer vs mouse in \cite{hu2018relation}), we focus on extracting features from a video for the same object instance.

%\vspace{-0.3cm}
\subsection{Object Detection In Videos}
%For video object detection, the main challenge is how to effectively exploit the temporal information for the same object instance from a video. Existing methods usually perform image-level or proposal-level feature aggregation to achieve this goal.
Existing methods usually perform image-level or proposal-level feature aggregation to exploit temporal information from a video for the same object instance.

\textbf{Image-level features across frames.}
There has been a line of research that boosts the single-frame detector by utilizing image-level features. D\&T  \cite{feichtenhofer2017detect} builds a dense correlation map between two feature maps of nearby video frames, and exploits instance track ids to learn frame-to-frame motion of bounding boxes. DFF \cite{zhu2017deep} firstly uses optical flow to propagate and align the features of selected keyframes to nearby non-keyframes in order to reduce redundant calculation and speed up the single-frame detector. FGFA \cite{zhu2017flow} utilizes optical flow to aggregate the nearby frames feature maps to target frame feature map in order to improve the single-frame detector. Furthermore, \cite{zhu2018towards} designs more advanced image-level feature propagation and keyframe selection mechanisms to achieve a better trade-off between accuracy and speed than FGFA. In order to avoid explicit optical flow computation, STSN \cite{bertasius2018object} proposes using deformable convolution to compute the offsets for feature alignment. \cite{wang2018non} proposes non-local module in order to completely drop the locality of estimated correspondences. However, the performance of these methods degrades quickly with longer time interval. In contrast to these methods, the proposed Temporal ROI Align doesn't perform image-level feature alignment and aggregation, and can take advantage of longer video length (even the entire video) than these methods.

\textbf{Proposal-level features across frames.}
Another line of research attempts to  aggregate the proposal-level features across frames. MANet \cite{wang2018fully} firstly utilizes optical flow to propagate current frame proposals to nearby frames, then aggregates the features of proposals from multi-frames. RDN \cite{deng2019relation} introduces the object relation module from still images into videos. It builds two stage relation modules to fully exploit the relationships among proposals. As mentioned above, the performance of these methods degrades quickly with longer time interval. Subsequent works attempt to overcome the shortcomings, i.e. previous works can't leverage long-range temporal information in videos. SELSA \cite{wu2019sequence} firstly generates proposals of multi-frames, and aggregates the high-level proposal features with each other. \cite{shvets2019leveraging} takes a further step. It develops a loss function which constraints the the feature aggregation conducted among the proposals belong to the same object instance. Although these methods perform high-level proposals' feature aggregation to utilize the temporal information in videos, they still adopt ROI Align operation to extract the ROI features for proposals, making the extracted ROI features lack temporal information from videos. In contrast to these methods, the proposed Temporal ROI Align can directly extract the temporal information from videos for proposals. Besides, the proposed operator can also be applied into these methods to further boost the performance. 
\vspace{-0.05cm}
\section{Temporal ROI Align}

\begin{figure}[t]
\centering
\includegraphics[width=8.45cm]{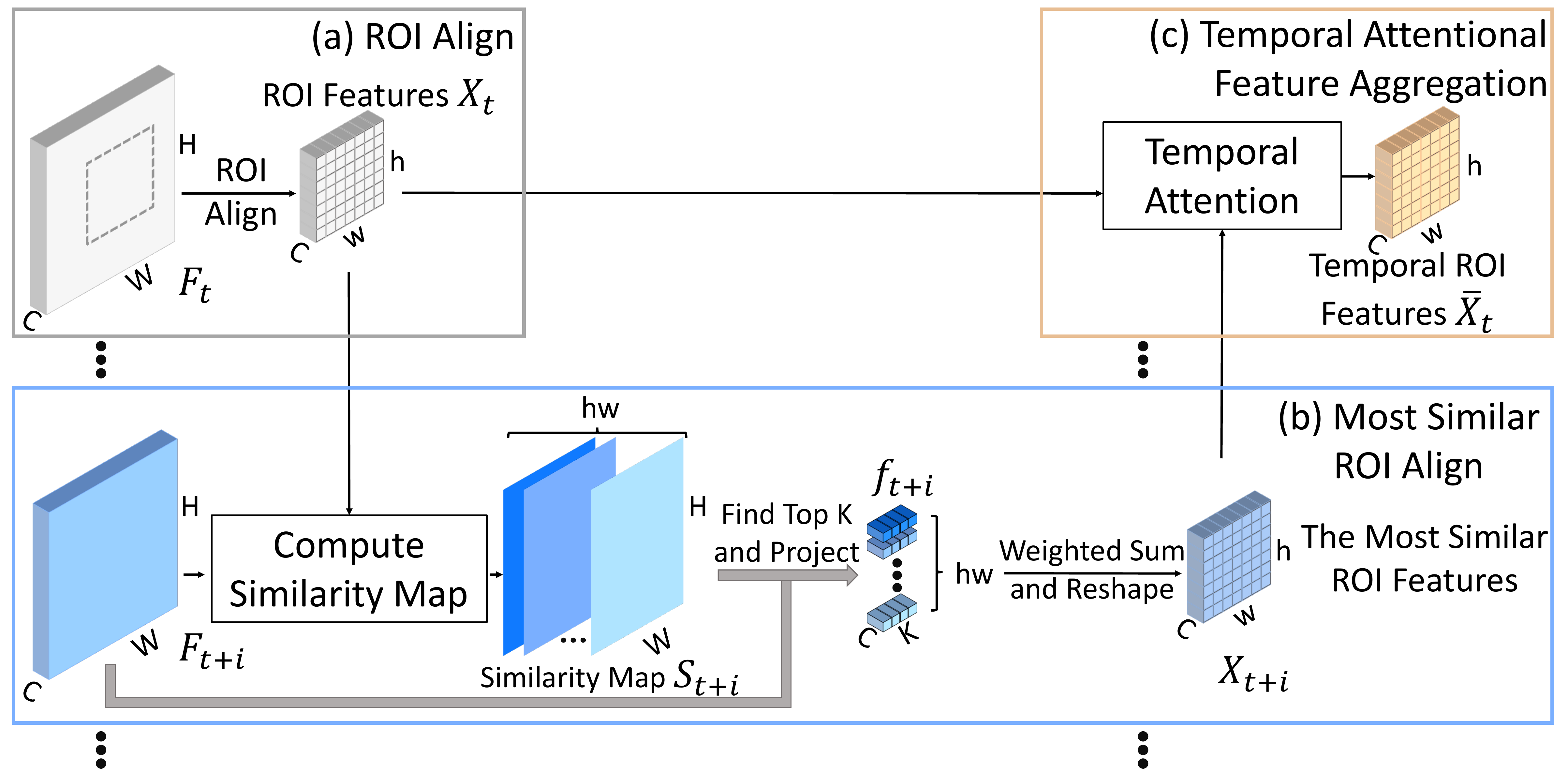}
\vspace{-0.6cm}
\caption{Temporal ROI Align. Firstly, the ROI features $X_t$ is extracted from $target$ frame feature map $F_t$ for target frame proposal in module (a). Then, Most Similar ROI Align (MS ROI Align) extracts the most similar ROI features $X_{t+i}$ from $support$ frame feature map $F_{t+i}$ for target frame proposal in module (b). Finally, Temporal Attentional Feature Aggregation (TAFA) performs temporal attention to aggregate $\{X_{t+i}\}_{i=-T/2}^{T/2}$ in module (c)
%Temporal ROI Align. There are three steps in Temporal ROI Align. Step 0 (not marked in the figure for convenience) extracts the ROI features $X_t$ for target frame proposals from target frame feature map $F_t$ by the conventional ROI Align operator. Step 1 takes the support frames feature maps $F_{t+i}$ and the ROI features $X_t$ as input. It aims at extracting the most similar ROI features $X_{t+i}$ from $F_{t+i}$ for target frame proposals based on the similarity maps $S_{t+i}$ between $X_t$ and $F_{t+i}$. $X_{t-i}$ is extracted in the same way with $F_{t-i}$ as support frame feature map. Step 2 takes $X_{t-i}, X_{t+i}$ and $X_t$ as input, and performs temporal attention to aggregate them in order to generate the final temporal ROI features $\overline{X}_t$. The temporal ROI features $\overline{X}_t$ not only contain the object features from target frame $t$, but also contain the same object instance features from support frames $t-i, t+i$
}
\label{fig:method}
\vspace{-0.25cm}
\end{figure}

The way to extract the features of proposals is crucial in videos. Existing methods usually adopt ROI Align operation to implement it. However, ROI Align is designed to extract the features of proposals from a single image features, making the extracted ROI features lack the temporal information for video tasks.

We propose Temporal ROI Align operator as an alternative to make the features of proposals contain temporal information of videos. Given a set of frames $\{I_{t+i}\}_{i=-T/2}^{T/2}$ from the same video, a set of feature maps $\{F_{t+i}\}_{i=-T/2}^{T/2}$ is extracted by a backbone network $g_{cnn}(\cdot\ ;\theta_{cnn})$:
\begin{equation}
F_{t+i} = g_{cnn}(I_{t+i}; \theta_{cnn})
\end{equation}
where $t$ denotes the index of target frame and $t+i$ $(i\neq0)$ denotes the index of support frames. $T$ is the total number of support frames. As shown in Fig. \ref{fig:method}, firstly, the ROI features $X_t$ is extracted for target frame proposals from $target$ frame feature map $F_t$ by the conventional ROI Align operator in module (a). Then, Most Similar ROI Align (MS ROI Align) focuses on extracting the most similar ROI features $X_{t+i}$ from $support$ frame feature map $F_{t+i}$ for target frame proposals in module (b). Specifically, similarity maps are calculated between $F_{t+i}$ and each spatial location of $X_t$. For each similarity map, we find the top K similarity scores as the most similar points, and project these points into $F_{t+i}$. Based on these points, the most similar features $f_{t+i}$  can be extracted from $F_{t+i}$. $f_{t+i}$ is weightedly summed by the normalized top K similarity scores to generate the most similar ROI features $X_{t+i}$. Finally, Temporal Attentional Feature Aggregation (TAFA) utilizes temporal attention to aggregate $\{X_{t+i}\}_{i=-T/2}^{T/2}$ in order to generate the final temporal ROI features $\overline{X}_t$ in module (c). The temporal ROI features $\overline{X}_t$ contain the object features from target frame and %the same object instance features from other 
support frames in a video.

%\vspace{-0.25cm}
%\subsection{Step 1: Extracting The Most Similar ROI Features}
\subsection{Most Similar ROI Align}
\begin{figure}[t]
\centering
\includegraphics[width=8.5cm]{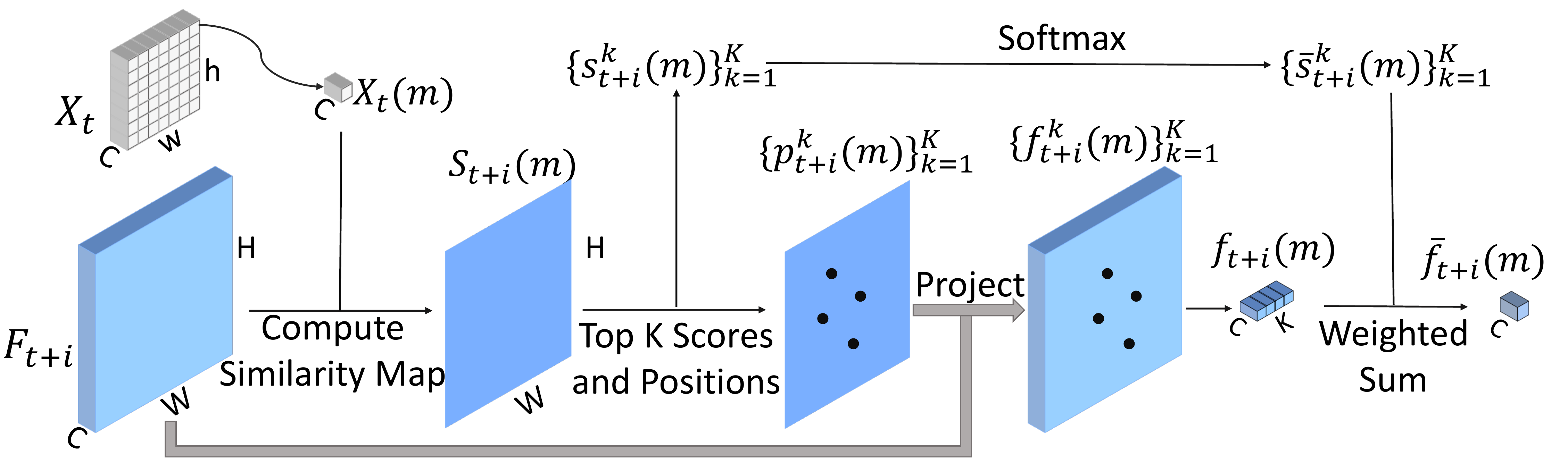}
\caption{The technical detailed illustration of MS ROI Align in the proposed Temporal ROI Align}
\label{fig:step1}
\vspace{-0.4cm}
\end{figure}
MS ROI Align aims at extracting the most similar ROI features from support frames feature maps for target frame proposals. Considering the features of an object instance are highly similar among different frames, we propose to implicitly extract the ROI features from support frames feature maps for target frame proposals based on feature similarity.

As shown in Fig. \ref{fig:step1}, the input is ROI features $X_t\in\mathbb{R}^{h\times w\times C}$ and the support frame feature map $F_{t+i}\in\mathbb{R}^{H\times W\times C}$. $h$, $w$ and $C$ denote the height, width and channel of $X_t$ respectively. $H$ and $W$ denote the height and width of $F_{t+i}$ respectively. One spatial location $X_t(m)$ is used to describe the technical details, where $m$ denotes the spatial location of $X_t$ and $m\in\{(1,1),(1,2), \cdots, (h,w)\}$. Other spatial locations of $X_t$ are performed in the same way.

%As shown in the Step 1 of Fig. \ref{fig:method}, we take the bottom Step 1 as example to describe the technical details. The top Step 1 is performed in the same way. The input is ROI features $X_t\in\mathbb{R}^{h\times w\times C}$ and the support frame feature map $F_{t+i}\in\mathbb{R}^{H\times W\times C}$. $t$ denotes target frame and $t+i$ denotes one of the support frames. $C$, $h$ and $w$ denote the channel, height and width of ROI features $X_t$ respectively. $H$ and $W$ denote the height and width of support frame feature map $F_{t+i}$ respectively.

Firstly both $X_t(m)$ and $F_{t+i}$ are L2 normalized along with the channel dimension to generate $\widehat{X}_t(m)$ and $\widehat{F}_{t+i}$.
%The normalization is important since the similarity maps should calculate the actual similarity between $X_t$ and $F_{t+i}$ rather than being dominated by feature magnitudes.
Then, cosine similarity map $S_{t+i}(m)\in\mathbb{R}^{H\times W}$ is calculated as follows:
\begin{equation}
\small
S_{t+i}(m) = S(\widehat{X}_t(m), \widehat{F}_{t+i}):\mathbb{R}^{1\times C} \otimes [\mathbb{R}^{(H\times W)\times C}]^T \rightarrow \mathbb{R}^{ H\times W}
\end{equation}
where $\otimes$ denotes matrix multiplication and $[\cdot]^T$ denotes matrix transposition.

Then, the top $K$ similarity scores $\{s_{t+i}^k(m)\}_{k=1}^{K}\in\mathbb{R}^{K}$ and corresponding spatial locations $\{p_{t+i}^k(m)\}_{k=1}^{K}\in\mathbb{R}^{K\times 2}$ are found from the similarity map $S_{t+i}(m)\in\mathbb{R}^{H\times W}$.
%\begin{equation}
%\{s_{t+i}^k(m)\}_{k=1}^{K}, \{p_{t+i}^k(m)\}_{k=1}^{K} = FindTopK(S_{t+i}(m))
%\end{equation}
%where $K$ is a hyperparameter.
The locations  $\{p_{t+i}^k(m)\}_{k=1}^{K}$ are projected into the support frame feature map $F_{t+i}$ in order to extract the most similar features  $\{f_{t+i}^k(m)\}_{k=1}^{K}$ for $X_t(m)$. $f_{t+i}(m)$ is defined as the set of $\{f_{t+i}^k(m)\}_{k=1}^{K}$ and $f_{t+i}(m)\in\mathbb{R}^{K\times C}$.
%\begin{equation}
%\{f_{t+i}^k(m)\}_{k=1}^{K} = Project(\{p_{t+i}^k(m)\}_{k=1}^{K}\rightarrow F_{t+i})
%\end{equation}
%where $f_{t+i}(m)$ is the set of $\{f_{t+i}^k(m)\}_{k=1}^{K}$ and $f_{t+i}(m)\in\mathbb{R}^{K\times C}$.

Finally, the similarity scores $\{s_{t+i}^k(m)\}_{k=1}^{K}$ are used to weight the $f_{t+i}(m)$:
\begin{equation}
\overline{s}_{t+i}^k(m) = \frac{\exp(s_{t+i}^k(m))}{\textstyle\sum_{k=1}^{K}\exp(s_{t+i}^k(m))}
\end{equation}

\begin{equation}
\overline{f}_{t+i}(m) = \textstyle\sum_{k=1}^{K}\overline{s}_{t+i}^k(m)f_{t+i}^k(m)
\end{equation}
where $\overline{s}_{t+i}^k(m)$ is the normalized weight of $s_{t+i}^k(m)$. $\overline{f}_{t+i}(m)\in\mathbb{R}^{C}$ is the weighted most similar features of $X_t(m)$.

Since there are total $h\times w$ spatial locations of ROI features $X_t$, the final most similar ROI features $X_{t+i}\in\mathbb{R}^{h\times w\times C}$ can be extracted by concatenating all $\overline{f}_{t+i}(m)$.
%\begin{equation}
%X_{t+i} = ConcatenateAndReshape(\overline{f}_{t+i}(1), \cdots, \overline{f}_{t+i}(hw))
%\end{equation}
The most similar ROI features $X_{t+i}$ can be regraded as the ROI features extracted from support frame feature map $F_{t+i}$ for the target frame proposals, since $X_{t+i}$ is the most similar features with $X_t$ in each spatial location.

%\subsection{Step 2: Aggregating These ROI Features}
%\vspace{-0.15cm}
\subsection{Temporal Attentional Feature Aggregation}
\begin{figure}[t]
\centering
\includegraphics[width=8.45cm]{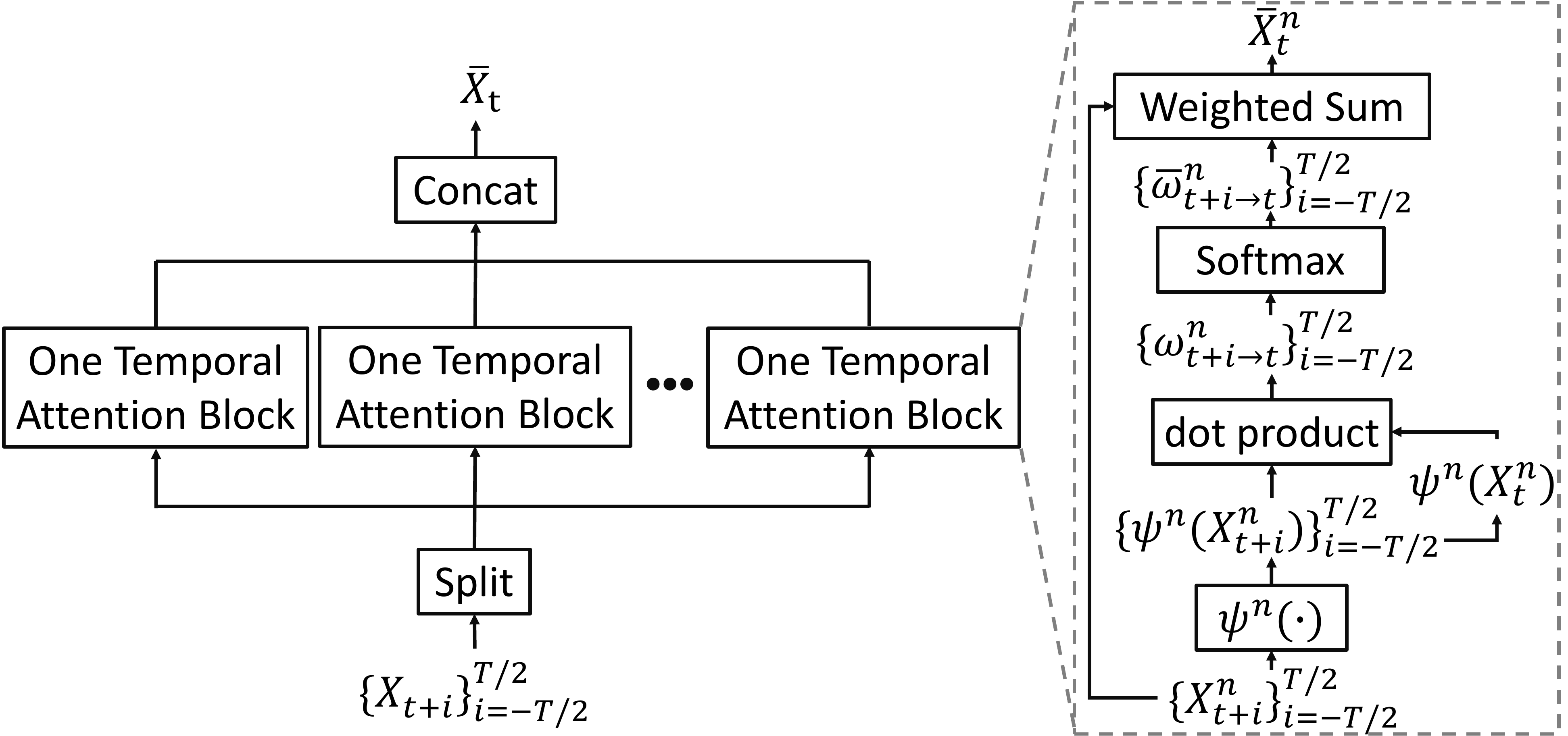}
%\vspace{-0.2cm}
\caption{Temporal Attentional Feature Aggregation in the proposed Temporal ROI Align}
\label{fig:step2}
\vspace{-0.05cm}
\end{figure}

In the MS ROI Align, we have already extracted the ROI features from target frame feature map and support frames feature maps. The remaining key challenge now becomes how to effectively aggregate these ROI features. 
%In some cases like object disappears in some support frames, the most similar ROI features may be "incorrect". Thus, it is natural to learn a group of temporal attention weights to aggregate them.
Since an object instance may be blurry in some frames and clear in other frames, it is natural to learn a group of temporal attention weights to aggregate them.
Besides, multi-head attention \cite{vaswani2017attention} allows model to jointly attend information from different representation subspaces at different channels. Therefore, we constitute multi temporal attention blocks to deal with different patterns in the temporal feature aggregation.
%Motivated by the multi-head attention in \cite{vaswani2017attention}, we constitute multi temporal attention blocks to aggregate these ROI features in order to better utilize them.

As shown in Fig. \ref{fig:step2}, the input is the set of ROI features $\{X_{t+i}\}_{i=-T/2}^{T/2}$. There are $N$ temporal attention blocks to aggregate $\{X_{t+i}\}_{i=-T/2}^{T/2}$. Firstly, $\{X_{t+i}\}_{i=-T/2}^{T/2}$ are split into N groups along the channel dimension:
%\vspace{-0.2cm}
\begin{equation}
X^n_{t+i} = X_{t+i}[:,:,(n-1)\frac{C}{N}:n\frac{C}{N}]
\end{equation}
where $X_{t+i}^n\in\mathbb{R}^{h\times w\times \frac{C}{N}}$ and $n\in\{1,2,\cdots, N \}$. Each $\{X_{t+i}^n\}_{i=-T/2}^{T/2}$ is used to generate one attention map:
\begin{equation}
w^n_{t+i\rightarrow t}(m)=\psi^n(X_{t+i}^n)(m)\cdot\psi^n(X_{t}^n)(m)
\end{equation}
\begin{equation}
\overline{w}^n_{t+i\rightarrow t}(m)=\frac{\exp(w^n_{t+i\rightarrow t}(m))}{\textstyle\sum_{i=-T/2}^{T/2}\exp(w^n_{t+i\rightarrow t}(m))}
\end{equation}
where $\psi^n(\cdot)$ is a tiny embedding network (network weights are not shared across different temporal attention blocks) and $\psi^n(X_{t+i}^n)\in\mathbb{R}^{h\times w\times \frac{C}{N}}$. $m$ denotes the spatial position of $\psi^n(X_{t+i}^n)$ and $\psi^n(X_{t+i}^n)(m)\in\mathbb{R}^{\frac{C}{N}}$. $w^n_{t+i\rightarrow t}\in\mathbb{R}^{h\times w}$ and $w^n_{t+i\rightarrow t}(m)$ is the attention weights from $t+i$ frame to $t$ frame of the $n$-th group attention map on spatial location $m$. $\overline{w}^n_{t+i\rightarrow t}(m)$ is the normalized attention weights across frames. Based on the normalized $n$-th group of temporal attention map $\{\overline{w}^n_{t+i\rightarrow t}\}_{i=-T/2}^{T/2}$, the $n$-th group of ROI features $\{X_{t+i}^n\}_{i=-T/2}^{T/2}$ are weightedly summed as follows:
\begin{equation}
\overline{X}^n_t(m)= \textstyle\sum_{i=-T/2}^{T/2}\overline{w}^n_{t+i\rightarrow t}(m)X^n_{t+i}(m)
\end{equation}
where $\overline{X}^n_t\in\mathbb{R}^{h\times w\times \frac{C}{N}}$. The final temporal ROI features $\overline{X}_t$ can be obtained by concatenating all $\overline{X}^n_t$ along the channel dimension and $\overline{X}_t\in\mathbb{R}^{h\times w\times C}$. The $\overline{X}_t$ is the same size as $X_t$, but it contains the temporal information of the same object instance in a video.
%\begin{equation}
%\overline{X}_t = Concatenate(\overline{X}^1_t, \cdots, \overline{X}^N_t)
%\end{equation}
%where $\overline{X}_t\in\mathbb{R}^{h\times w\times C}$. The $\overline{X}_t$ is the same size as $X_t$, but it contains the temporal information of the same object instance in a video compared with $X_t$.

In some cases like object disappears in some support frames, similar features of MS ROI Align from these supporting frames may be "incorrect". The proposed TAFA can alleviate the issue by the attention weight. 
%The proposed TAFA can alleviate the adverse effect of "incorrect" most similar ROI features by the attention weight.
The “incorrect” features can be suppressed by the softmax operator if “correct” features (i.e. object features) can be extracted from other support frames. Overall, the MS ROI Align is responsible for extracting the "correct" object features as far as possible. If it fails, the TAFA will minimize the adverse influence of the "incorrect" features by the attention weight. The MS ROI Align with TAFA guarantees that the proposed Temporal ROI Align works on most cases.

Temporal ROI Align is designed as a general operator to extract better features for objects, and it can also work under some special cases. For example, when an object in target frame is partial occlusion, Temporal ROI Align can still extract the object features from support frames for most spatial locations without occlusion. Therefore, the visible parts are dominant and features from these locations can still get enhanced.
%\vspace{-0.15cm}
\section{Experiments On VID}
%Section \ref{exp_dataset} briefly introduces the dataset and evaluation metrics used for VID. The implementation details are described in Section \ref{exp_implement_details}. We next conduct ablation experiments in Section \ref{exp_ablation} and show the performance comparison with other methods in Section \ref{exp_sota}. Section \ref{exp_visualization} explains why the proposed Temporal ROI Align can boost the performance of video detectors by visualization.

%%\vspace{-0.25cm}
\subsection{DataSet And Evaluation Metric}
\label{exp_dataset}

Experiments are carried out on the ImageNet VID dataset \cite{russakovsky2015imagenet} which contains 30 object categories for video object detection.
%which is a prevalent large-scale benchmark for video object detection. 
There are total 3862 video snippets in the training set and 555 video snippets in the validation set. 
%All the frames are fully annotated with bounding boxes across 30 object categories.
%The models are trained with a mixture of ImageNet VID and DET datasets (using 30 VID classes, the subset of 200 DET classes). For fair comparison, we use the split provided in FGFA \cite{zhu2017flow}. At most 15 frames are subsampled from each video, and the DET:VID balance is approximately 1:1.
The mAP@IoU=0.5 is reported on the validation set.

%\vspace{-0.15cm}
\subsection{Implementation Details}
\label{exp_implement_details}

\noindent\textbf{Backbone Network.}
The ResNet-101 \cite{he2016deep} (R101) is used as the backbone network $g_{cnn}(\cdot\ ;\theta_{cnn})$ for ablation studies. ResNeXt-101-64x4d \cite{xie2017aggregated} (X101) is also used for the final results. The stride of the first conv block in the $conv5$ stage of convolutional layers is modified from 2 to 1 in order to enlarge the resolution of feature maps. As such, the effective stride in the stage is changed from 32 pixels to 16 pixels. All the $3\times 3$ conv layers in the stage are modified by the dilated convolutions to compensate the receptive fields.

\noindent\textbf{Region Proposal Network.}
RPN is placed on the output of $conv4$ to generate proposals. There are a total of 12 anchors with 4 scales $\{64^2, 128^2, 256^2, 512^2\}$ and 3 aspect ratios $\{1:2, 1:1, 2:1\}$. 300 proposals are produced on each image.

\noindent\textbf{Temporal ROI Align.}
We replace the conventional ROI Align operator with the proposed Temporal ROI Align operator. The Temporal ROI Align is applied on the ouput of $conv5$. The spatial size $h, w$ of the ROI features are set to $7$. For MS ROI Align, there are a total of 49 similarity maps needed to be calculated for each proposal, and the top 4 ($K=4$) similarity scores and corresponding spatial locations are selected for each similarity map. For TAFA, a total of 4 ($N=4$) temporal attention blocks are used to aggregate the ROI features and these most similar ROI features, and each $\psi^n(\cdot)$ is a $3\times 3$ convolution layer.

\noindent\textbf{Detection Network.}
Two 1024-d fully connected layers are applied upon the temporal ROI features followed by classification and bounding box regression.

\noindent\textbf{Training and Testing Details.}
The backbone networks are initialized with ImageNet pre-trained weights. A total of 6 epochs of SGD training is performed with a total batch size of 16 on 16 GPUs. The initial learning rate is 0.02 and is divided by 10 at the 4-th and 6-th epoch. The models are trained with a mixture of ImageNet VID and ImageNet DET datasets \cite{russakovsky2015imagenet} (using 30 VID classes and the overlapping 30 of 200 DET classes). For fair comparison, we use the split provided in FGFA \cite{zhu2017flow}. At most 15 frames are subsampled from each video, and the DET:VID balance is approximately 1:1. For training, one training frame is sampled along with two random frames from the same video. For inference, $T$ frames (support frames) from the same video are sampled along with the inference frame (target frame). Following STSN [1], if support frames are beyond the video start/end, we copy the first/last frame of the video. NMS with a threshold of 0.5 is adopted to suppress reduplicate detection boxes. In both training and inference, the images are resized to a shorter side of 600 pixels.

%\vspace{-0.15cm}
\subsection{Ablation Study}
\label{exp_ablation}
\begin{figure}[t]
\centering
\includegraphics[width=8.425cm]{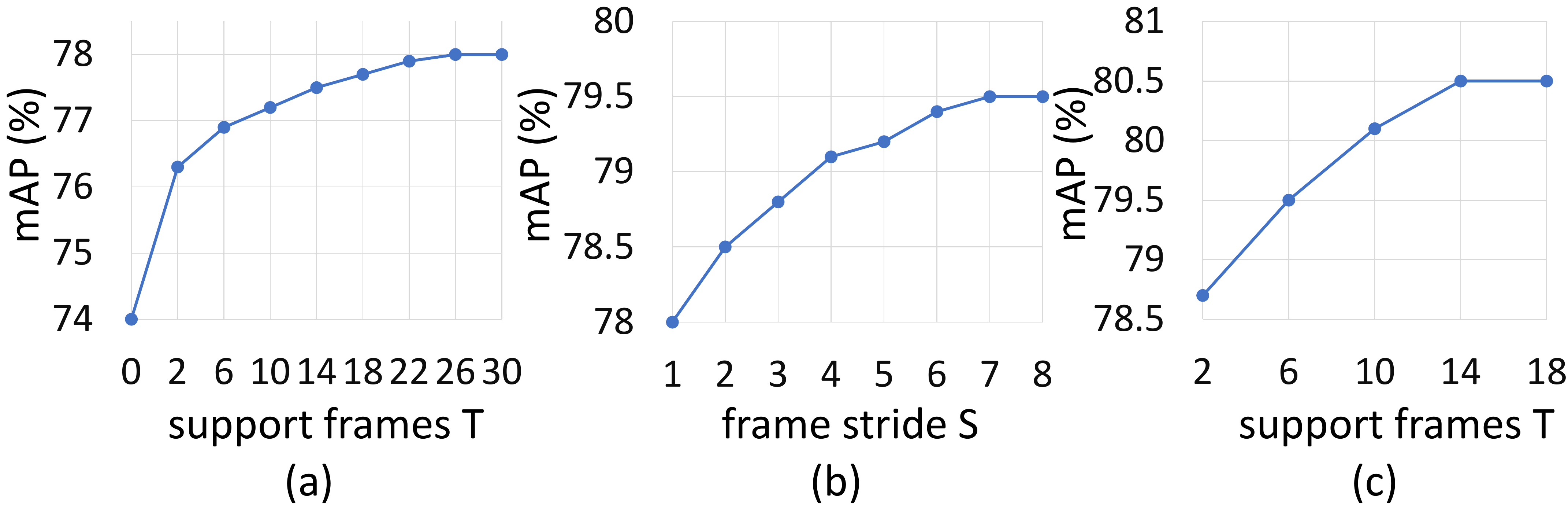}
\caption{Ablation for different support frame sampling strategies. (a) The effect of different number of support frames $T$. The frame stride $S$ is fixed to 1. (b) The effect of different sampling frame stride $S$. The number of support frames $T$ is fixed to $26$. (c) The effect of different number of support frames $T$. The support frames $T$ are uniformly  sampled from the entire video}
\label{fig:support_frame_sample_strategy}
\vspace{-0.25cm}
\end{figure}

\noindent\textbf{Support Frame Sampling Strategies.}
The sampling strategy for support frame is crucial for video object detection.
%Using more consecutive support frames usually brings more performance gains. \cite{shvets2019leveraging} and \cite{wu2019sequence} also find that sampling support frames with a uniform stride can further improve the performance by leveraging long-range temporal information from videos. The proposed Temporal ROI Align can also inherit the advantage of \cite{shvets2019leveraging} and \cite{wu2019sequence}.
Here, we conduct experiments to analyse the effect of the number of support frames $T$ and sampling frame strides $S$.
Specifically, by using a sampling frame stride $S$, one frame in every $S$ frames is used for testing rather than consecutive frames.

%Firstly, the $S$ is fixed to 1 and we vary the number of $T$, i.e., the support frames are consecutive frames.
Firstly, we fix the stride $S$ to 1 and vary the number of $T$ to analyse the effect of the number of consecutive support frames. As shown in Fig. \ref{fig:support_frame_sample_strategy} (a), the single-frame detector baseline, which only uses the conventional ROI Align to extract features from target frame feature map, achieves 74.0 mAP without any support frames. As the number of support frames increases, the performance can be improved consistently and finally stabilized at 78.0 mAP after $T=26$. It means that using more consecutive support frames usually can bring more performance gains.
%The single-frame detector baseline only uses the conventional ROI Align to extract features from target frame feature map. As shown in Fig. \ref{fig:support_frame_sample_strategy} (a), the single-frame detector baseline is 74.0 mAP without support frames. With more and more support frames, the performance increases consistently. When $T=26$, the performance seems stabilize with 78.0 mAP.

Then, we fix the number of support frames $T$ to 26 and vary the number of $S$ to explore the effect of the fixed sampling stride. As shown in Fig. \ref{fig:support_frame_sample_strategy} (b), the performance increases steadily with $S$ raising and finally stabilizes on 79.5 mAP after $S=7$.
It indicates that sampling support frames with a larger stride can further improve the performance since longer-range temporal information can be leveraged, which also coincides the results in \cite{shvets2019leveraging} and \cite{wu2019sequence}.

Finally, to further utilize the whole video information, we uniformly sample the support frames from the entire video, i.e. the sampling stride $S$ is adaptive to the length of the video according to the number of the support frames $T$. Fig. \ref{fig:support_frame_sample_strategy} (c) shows the performance with different values of $T$.
%However, different with Fig. \ref{fig:support_frame_sample_strategy} (a), the support frames are uniformly sampled from the entire video. As seen in Fig. \ref{fig:support_frame_sample_strategy} (c),
We can see that with 2 support frames that are uniformly sampled from the entire video, the performance already achieves 78.7, which outperforms the 26 consecutive support frames by 0.7 point. With 6 support frames that are uniformly sampled from the entire video, the performance can achieve the same results with 26 support frames that are sampled by the fixed sampling frame stride 8. The performance keeps increasing as $T$ rises and finally stabilizes at 80.5 mAP after $T=14$.
These results demonstrate that the proposed Temporal ROI Align can utilize long-range temporal information from videos. The uniform sampling strategy with $T=14$ is used as the default setting in the following experiments.
%The experiments show that the proposed Temporal ROI Align can utilize long-range temporal information from videos. This is the default test setting in the following experiments.

\noindent\textbf{The Effectiveness of MS ROI Align and TAFA.}
\begin{table}[t]
\begin{center}
\begin{tabular}{c|c}
  \hline
  \hline
  Method & mAP(\%) \\
  \hline
  \hline
  Single-frame Detector & 74.0 \\
  \hline
  MD\ ROI\ Align + Averaging & 68.5\\
  MD\ ROI\ Align + TAFA & 76.6\\
  MS\ ROI\ Align + Averaging & 78.5\\
  MS\ ROI\ Align + TAFA & \textbf{80.5}\\
  \hline
  \hline
\end{tabular}
\end{center}
\vspace{-0.3cm}
\caption{The Effectiveness of MS ROI Align and TAFA. "MD ROI Align" denotes that the MS ROI Align is replaced by Modulated Deformable ROI Align. "Averaging" denotes the TAFA is replaced by simply averaging}
\label{t:temporal_roi}
%\vspace{-0.0cm}
\end{table}
%Several experiments are conducted to show the effectiveness of MS ROI Align and TAFA in the proposed Temporal ROI Align.
MS ROI Align in the proposed Temporal ROI Align is designed to find the location corresponding to the proposal of target frame from the support frame feature maps. Inspired by STSN \cite{bertasius2018object} that utilizes deformable convolutions to calculate the offsets used for image feature calibration, we replace the ROI Align with the recently proposed Modulated Deformable ROI Align \cite{zhu2019deformable} (MD ROI Align) to show the effectiveness of MS ROI Align. The MD ROI Align can calculate the offsets between target frame and support frames in order to extract the ROI features from support frames feature maps for target frame proposals. As shown in Table \ref{t:temporal_roi}, "MD ROI Align + TAFA" achieves 76.6 mAP, which only exceeds the single-frame baseline by 2.6 points. The reason behind this limited improvement lies in that deformable operation can't learn the effective offsets when the frame stride is large, which is also shown in the results of STSN \cite{bertasius2018object} where the performance degrades from 78.9 to 77.9 when the frame stride changes from 1 to 4. Compared to "MS ROI Align + TAFA", the performance of "MD ROI Align + TAFA" drops around 4 points, which demonstrates the superiority of the designed MS ROI Align.

To demonstrate the effectiveness of TAFA for feature aggregation, we also conduct a simple baseline where the ROI features from target frame and support frames are directly averaged. The experiment results are shown in Table \ref{t:temporal_roi}. We can see that "MS ROI Align+ Averaging" only achieves 78.5 mAP, which is lower than "MS ROI Align+ TAFA" by 2 points. This demonstrates that the temporal attention design in TAFA can learn to put more weights on better ROI features.

\noindent\textbf{The Effectiveness of Temporal ROI Align.}
\begin{table}[t]
\begin{center}
\begin{tabular}{c|c}
  \hline
  \hline
  Method & mAP(\%) \\
  \hline
  \hline
  Single-frame Detector & 74.0 \\
  \hline
  ROI Align + Non-Local & 77.9 \\
  ROI Align + Non-Local (Top K) & 78.4 \\
  Temporal ROI Align & \textbf{80.5}\\
  \hline
  \hline
\end{tabular}
\end{center}
\vspace{-0.2cm}
\caption{Performance comparison with Non-local and Temporal ROI Align}
\label{t:non-local}
\vspace{-0.2cm}
\end{table}
Since Non-Local \cite{wang2018non} can also extract the temporal information from support frames, we also attempt to integrate ROI Align with Non-Local  module in order to show the superiority of Temporal ROI Align.  As shown in Table \ref{t:non-local}, the mAP is only 77.9 when applying Non-Local after ROI Align. Non-local inevitably aggregates background features into the ROI features, since the attention weights of Non-Local are computed from all spatial locations in  support frames. This leads to the inferior mAP of ROI Align + Non-Local. We also modify the original Non-Local by picking the Top K attention weights to eliminate the background features as far as possible. However, the result (78.4) is still lower than the proposed Temporal ROI Align (80.5) by 2.1 points. This shows the superior performance of the proposed Temporal ROI Align.

\noindent\textbf{Experiments on Hyperparameters.}
\begin{table}[t]
\begin{center}
\begin{tabular}{c|c|c|c|c|c|c}
  \hline
  \hline
  \#$K$ & 1 & 2 & 3 & 4 & 5 & 6\\
  \hline
  mAP(\%) &79.3 & 79.8 & 80.2 & \textbf{80.5} & 80.4 & 80.2\\
  \hline
  \hline
\end{tabular}
\end{center}
\vspace{-0.2cm}
\caption{The effect of different the most similar K locations in MS ROI Align of Temporal ROI Align}
\label{t:topk}
%\vspace{-0.1cm}
\end{table}
We analyse the effect of two hyperparameters $K$ and $N$, standing for the number of the most similar locations on each similarity map and the number of the temporal attention blocks respectively.

Table \ref{t:topk} shows the effect of different $K$. With the $K$ increasing from 1 to 4, the performance increases from 79.3 mAP to 80.5 mAP. This shows that the extracted most similar ROI features can benefit from more sampling location $K$. However, the performance seems to have a downward trend when $K$ is larger than 4, since using large $K$ may find the features of unrelated location, especially for small objects.
%probably because the extracted most similar ROI features suffer from the features of unrelated location, especially for small objects.
Therefore, $K$ is chosen to 4 as default setting.

\begin{table}[t]
\begin{center}
\begin{tabular}{c|c|c|c|c|c|c}
  \hline
  \hline
  \#$N$ & 1 & 2 & 4 & 8 & 16 & 32\\
  \hline
  mAP(\%) &79.5 & 80.1 & \textbf{80.5}& \textbf{80.5} & 80.4 & 80.3\\
  \hline
  \hline
\end{tabular}
\end{center}
\vspace{-0.2cm}
\caption{The effect of different number of the temporal attention blocks $N$ in TAFA of Temporal ROI Align}
\label{t:attention_maps}
\vspace{-0.2cm}
\end{table}
Table \ref{t:attention_maps} shows the effect of using different number of the temporal attention blocks $N$ in TAFA. With the $N$ increasing from 1 to 4, the performance increases from 79.5 mAP to 80.5 mAP. This shows that using more temporal attention blocks can improve the accuracy. The performance saturates when $N$ is bigger than 4. Therefore, $N$ is chosen to 4 as default setting.

%\vspace{-0.25cm}
\subsection{Comparison With State-of-the-art Methods}
\begin{table}[t]
\begin{center}
\begin{tabular}{c|c|c}
  \hline
  \hline
  Method & Backbone & mAP(\%) \\
  \hline
  \hline
  SELSA \cite{wu2019sequence} & R101& 80.25 \\
  \hline
  SELSA-$ReIm$ & R101& 80.3 \\
  \makecell[c]{SELSA-$ReIm$ + TROI} & R101 & \textbf{82.0} \\
  \hline
  \hline
  SELSA \cite{wu2019sequence} & X101 & 83.11 \\
  \hline
  SELSA-$ReIm$ & X101 & 83.0 \\
  \makecell[c]{SELSA-$ReIm$ + TROI} & X101 & \textbf{84.3} \\
  \hline
  \hline
\end{tabular}
\end{center}
\vspace{-0.2cm}
\caption{Applying the Temporal ROI Align (TROI) into SELSA video detector. Note that the SELSA-$ReIm$ denotes the reimplementation of SELSA}
\label{t:applied_selsa}
%\vspace{-0.1cm}
\end{table}
The proposed Temporal ROI Align offers an alternative for the conventional ROI Align operator. Therefore, it is easy to be  implemented into other video detectors. We reproduce the newly proposed SELSA \cite{wu2019sequence} due to its effectiveness, simpleness and also the state-of-the-art performance.
As shown in Table \ref{t:applied_selsa}, SELSA-$ReIm$ denotes the reimplementation of SELSA, and it basically achieves the same performance with SELSA \cite{wu2019sequence} under ResNet-101 and ResNeXt-101 backbones. When replacing ROI Align by the proposed Temporal ROI Align, the performance further improves by 1.7 points and 1.3 point respectively, which demonstrates the flexibility of the proposed operator.

\label{exp_sota}
\begin{table}[t]
\small
\begin{center}
\begin{tabular}{c|c|c}
  \hline
  \hline
  Method & Backbone & mAP(\%) \\
  \hline
  \hline
  %FGFA$^\triangle$ \cite{zhu2017flow} & R101 & 78.4 \\
  FGFA \cite{zhu2017flow} & R101 & 78.4 \\
  D\&T \cite{feichtenhofer2017detect} & R101 & 80.0 \\
  PLSA \cite{guo2019progressive} & R101+DCN & 80.0 \\
  %STSN$^\triangle$ \cite{bertasius2018object} & R101 + DCN & 80.4 \\
  SELSA \cite{wu2019sequence} &R101 & 80.25 \\
  Leveraging \cite{shvets2019leveraging} & R101-FPN & 81.0 \\
  %PLSA$^\triangle$ \cite{guo2019progressive} & R101+DCN & 81.4 \\
  RDN \cite{deng2019relation} & R101 & 81.8 \\
  %SELSA$*$ \cite{wu2019sequence} & R101 & 82.69 \\
  MEGA \cite{Chen_2020_CVPR} & R101 & 82.9 \\
  %HVR-Net \cite{han2020mining} & R101 & 83.2 \\
  %RDN$^\triangle$ \cite{deng2019relation} & R101 & 83.8 \\
  %HVR-Net$^\triangle$ \cite{han2020mining} & R101 & 83.8 \\
  %MEGA$^\triangle$ \cite{Chen_2020_CVPR} & R101 & 84.5 \\
  \hline
  \textbf{\makecell[c]{Single-frame Detector + TROI}} & R101 & \textbf{80.5} \\
  \textbf{\makecell[c]{SELSA \cite{wu2019sequence} + TROI}} & R101 & \textbf{82.0} \\
  \hline
  \hline
  D\&T \cite{feichtenhofer2017detect} & X101 & 81.6 \\
  SELSA \cite{wu2019sequence} &X101 & 83.11 \\
  RDN\cite{deng2019relation} & X101 & 83.2 \\
  Leveraging \cite{shvets2019leveraging} & X101-FPN & 84.1 \\
  MEGA \cite{Chen_2020_CVPR} & X101 & 84.1 \\
  %SELSA$*$ \cite{wu2019sequence} & X101 & 84.30 \\
  %RDN$^\triangle$ \cite{deng2019relation} & X101 & 84.7 \\
  %HVR-Net \cite{han2020mining} & X101 & 84.8 \\
  MEGA$^\triangle$ \cite{Chen_2020_CVPR} & X101 & 85.4 \\
  %HVR-Net$^\triangle$ \cite{han2020mining} & X101 & 85.5 \\
  \hline
  \textbf{\makecell[c]{Single-frame Detector + TROI} } & X101 & \textbf{82.3} \\
  \textbf{\makecell[c]{SELSA \cite{wu2019sequence} + TROI}} & X101 & \textbf{84.3} \\
  \hline
  \hline
\end{tabular}
\end{center}
\vspace{-0.2cm}
\caption{Performance comparison with other state-of-the-art models on the ImageNet VID validation set. $^\triangle$ denotes using heavy video post-processing methods like Seq-NMS \cite{han2016seq}. TROI denotes the proposed Temporal ROI Align 
%$^\triangle$ denotes using heavy video post-processing methods, such as Seq-NMS \cite{han2016seq}, BLR \cite{deng2019relation}. $*$ denotes using complicated data augmentation including photometric distortion, random expand and random crop in \cite{wu2019sequence}
}
\label{t:comparison_with_sota}
\vspace{-0.1cm}
\end{table}
Table \ref{t:comparison_with_sota} summarizes the performance of the proposed method and other state-of-the-art models 
%without post-processing 
on the ImageNet VID validation set.
Without bells and whistles, the proposed Temporal ROI Align can achieve 80.5 mAP with ResNet-101 backbone when applied to single-frame detector baseline. When applied to SELSA \cite{wu2019sequence}, the performance further improves to 82.0 mAP.
%with a speed of 9.0 FPS on a Tesla V100 GPU. 
%which is the best one compared to existing methods that adopt the same ResNet-101 backbone and do not use any post-processing techniques and complicated data augmentation either.
%The performance of ours is also competitive with SELSA$*$ (82.69) and RDN$^\triangle$ (83.8), despite SELSA$*$ utilizes very complicated data augmentation which includes photometric distortion, random expand and random crop, and RDN$^\triangle$ utilizes heavy video post-processing method BLR \cite{deng2019relation}. The complicated data augmentation brings 2.44 points performance gains for SELSA (from 80.25 to 82.69), and the BLR brings 2 points performance gains for RDN (from 81.8 to 83.8). Besides, our method uses less support frames (14 frames) than SELSA$*$ (21 frames) and RDN$^\triangle$ (36 frames). 
%SELSA$*$ achieves better performance (82.69) than ours by utilizing complicated data augmentation which includes photometric distortion, random expand and random crop. The complicated data augmentation brings 2.44 points performance gains for SELSA (from 80.25 to 82.69). RDN$^\triangle$ achieves better performance (83.8) than ours by utilizing heavy video post-processing method BLR \cite{deng2019relation}. The BLR brings 2 points performance gains for RDN (from 81.8 to 83.8). Besides, our method uses less support frames (14 frames) than SELSA$*$ (21 frames) and RDN$^\triangle$ (36 frames).
Furthermore, by using a stronger ResNeXt-101 backbone, Temporal ROI Align applied to SELSA achieves 84.3 mAP, which is the top one among existing methods that also adopt ResNeXt-101 backbone and do not use any post-processing techniques. Besides, our method uses less support frames (14 frames) than MEGA (35 frames), and  the pipeline of our method is more simple than MEGA.
%which means less computation overhead.
%The performance is comparable with MEGA (84.1) and HVR-Net (84.8).
%, while the pipeline of our method is more simple than RDN$^\triangle$.
%However, our method uses less support frames (14 frames) than MEGA (35 frames) and HVR-Net (31 frames), which means less computation overhead.

%\vspace{-0.25cm}
\subsection{Additional Experiments On EPIC KITCHENS}
ImageNet VID dataset falls short in the density and diversity of objects. Therefore we evaluate Temporal ROI Align on the EPIC KITCHENS dataset \cite{damen2018scaling}. EPIC contains 290 classes and is far more complex and challenging. 272 video sequences captured in 28 kitchens are used for training. 106 sequences collected in the same 28 kitchens (S1) and 54 sequences collected in other 4 unseen kitchens (S2) are used for evaluation. %Videos are annotated in 1s interval.
\begin{table}[t]
\begin{center}
\small
\begin{tabular}{c|c|c}
  \hline
  \hline
  Method & mAP@0.5 (S1) & mAP@0.5 (S2)\\
  \hline
  \hline
  SELSA \cite{wu2019sequence} & 38.0 & 34.8\\
  \hline
  SELSA-$ReIm$ & 38.8 & 36.7\\
  SELSA-$ReIm$ + TROI & \textbf{42.2} & \textbf{39.6}\\
  \hline
  \hline
\end{tabular}
\end{center}
\vspace{-0.2cm}
\caption{Performance comparison on EPIC KITCHENS test set. S1 and S2 indicate Seen and Unseen splits, respectively. SELSA-$ReIm$ denotes the reimplementation of SELSA. TROI denotes Temporal ROI Align}
\label{t:epic}
%\vspace{-0.0cm}
\end{table}

The results is shown in Table \ref{t:epic}. Note that the performance of SELSA-$ReIm$ (38.8 and 36.7) is higher than SELSA \cite{wu2019sequence} (38.0 and 34.8). When SELSA-$ReIm$ is equipped with the proposed Temporal ROI Align, the mAP@0.5 further improves to 42.2 and 39.6 by 3.4 and 2.9 points, respectively. This demonstrates that Temporal ROI Align can bring more performance gains on more complex video objection dataset compared with ImageNet VID.

\section{Extension: Experiments On VIS}
%\vspace{-0.1cm}
\begin{table}[t]
\small
\begin{center}
\begin{tabular}{c|c|c|c|c}
  \hline
  \hline
  Method & Backbone & AP & AP$_{50}$ & AP$_{75}$ \\
  \hline
  \hline
  MT R-CNN & R50-FPN & 30.3 & 51.1 & 32.6 \\
  \makecell[c]{MT-CNN + TROI} & R50-FPN & \textbf{33.5} & \textbf{57.0} & \textbf{36.6} \\
  \hline
  MT R-CNN & X101-FPN & 34.9 & 58.8 & 36.5 \\
  \makecell[c]{MT R-CNN + TROI} & X101-FPN & \textbf{38.0} & \textbf{63.3} & \textbf{40.3} \\
  \hline
  \hline
\end{tabular}
\end{center}
\vspace{-0.2cm}
\caption{Applying the Temporal ROI Align to MaskTrack R-CNN (MT R-CNN) in VIS. R50 denotes ResNet-50. AP denotes mask AP which follows the COCO evaluation metric to use 10 IoU thresholds from 50\% to 95\% at step 5\%}
\label{t:vis}
\vspace{-0.1cm}
\end{table}
We also investigate Temporal ROI Align on video instance segmentation (VIS). VIS
%is a newly proposed task in \cite{yang2019video}, which
aims at simultaneously detecting, segmenting and tracking object instances in a video sequence. The dataset of VIS is YouTube-VIS \cite{yang2019video} which contains 40 object categories. It contains 2238 training videos, 302 validation videos, and 343 test videos.
%Each video is annotated with per-pixel segmentation, bounding box, category, and instance labels.
%There are 40 object categories in the dataset. 
The training is performed on the training videos. Since the evaluation on the test set is currently closed, the evaluation is performed on the validation set.
%The evaluation metric is mask AP which follows the COCO evaluation metric to use 10 IoU thresholds from 50\% to 95\% at step 5\%.

%A video instance segmentation method, named MaskTrack R-CNN, is proposed in \cite{yang2019video}, which is
%\cite{yang2019video} also proposes a method, named as MaskTrack R-CNN, for YouTube-VIS. MaskTrack R-CNN is

MaskTrack R-CNN \cite{yang2019video} is based on Mask R-CNN, and builds another branch (track head) to link the same object instances between two frames. MaskTrack R-CNN uses the conventional ROI Align to extract features from single-frame for proposals, and we replace the ROI Align with the proposed Temporal ROI Align.
%We evaluate a standard baseline of ResNet-50-FPN and a higher baseline of ResNeXt-101-FPN using the default training setting for YouTube-VIS.
The results are shown in Table \ref{t:vis}. We can see that the Temporal ROI Align consistently improves ResNet-50-FPN and ResNeXt-101-FPN baselines on all metrics involving AP, AP$_{50}$ and AP$_{75}$ (e.g. 3.2 points in AP of ResNet-50-FPN and 3.1 points in AP of ResNeXt-101-FPN), which further demonstrates the flexibility of  proposed Temporal ROI Align.

\section{Visualization and Analysis}
\begin{figure*}
\centering
\includegraphics[width=12cm]{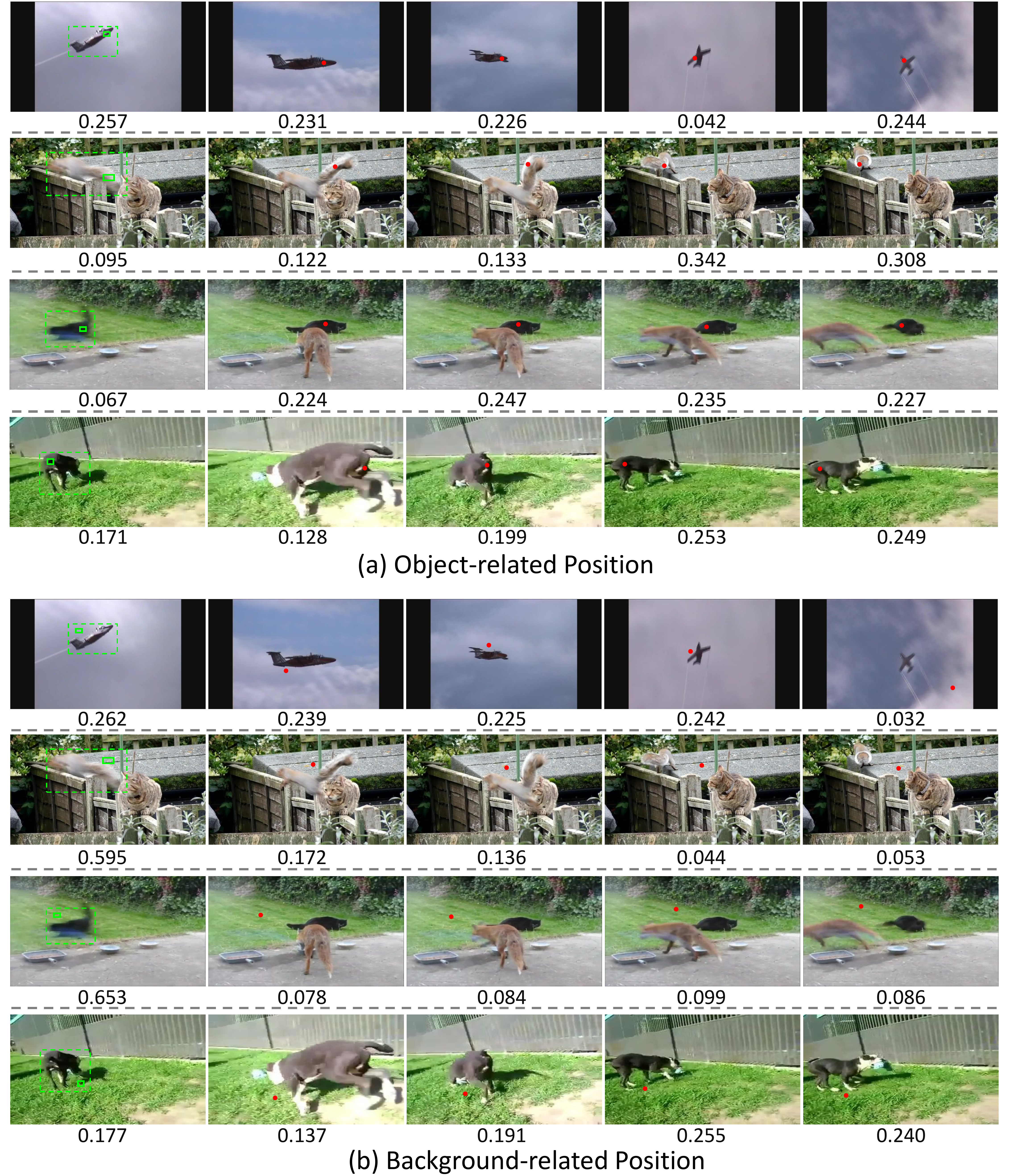}
\caption{An illustration of the most similar positions found by MS ROI Align and the corresponding averaged attention weights across different temporal attention blocks in TAFA (zoom in for a better view). The support frames $T$ is set to 4 during inference for visualization. The first column is the target frames and other columns are support frames. The dash green box is a proposal in target frames.
The solid green box is the region projected into the target frame from the object-related position (a) or background-related position (b) in the $7\times 7$ ROI features of the proposal. The red points in support frames are the most similar positions corresponding to the object-related position (a) or background-related position (b) of the $7\times 7$ ROI features in target frame. Only the top 1 similar position rather than top 4 similar  positions is visualized for a clear visualization. The floating numbers below each row of images are the averaged attention weights in TAFA corresponding to the object-related position (a) or background-related position (b) in the $7\times 7$ ROI features of the proposal}
\label{fig:supple_visualization}
\end{figure*}

We visualize the most similar positions found by MS ROI Align and the corresponding averaged attention weights across different temporal attention blocks in TAFA.

Fig. \ref{fig:supple_visualization} (a) shows the object-related position %(not only the center position in the manuscript) 
in the $7\times 7$ ROI features of the proposal. We can see that the red points are localized in the region of the same object instance, and the proposed TAFA can learn reasonable attention weights. For example, in the third row of Fig. \ref{fig:supple_visualization} (a), the objects are hard to be classified in the target frames due to motion blur. However, the object is clear in some support frames (e.g. the 3-rd row, 3-rd column in Fig. \ref{fig:supple_visualization} (a)) and the MS ROI Align successfully locates the object-related points (red points). The attention weights of points belonging to clear object (e.g. 0.247 in the 3-rd row, 3-rd column of Fig. \ref{fig:supple_visualization} (a)) are also larger than attention weights of points belonging to blurry object (e.g. 0.067 in the 3-rd row, 1-st column of Fig. \ref{fig:supple_visualization} (a)). After the most similar features from red points in the support frames being integrated with the original ROI features by the reasonable attention weights in TAFA, it is much easier for the classifier to make the correct classification.

Fig. \ref{fig:supple_visualization} (b) shows the background-related position in the $7\times 7$ ROI features of the proposal. We can see that most red points are localized in the surrounding region of the same object instance (e.g. the 1-rd row, 3-rd column in Fig. \ref{fig:supple_visualization} (b)), which can be viewed as the context information of the object instance. If the red points are localized in the background region which is far away from object instances (e.g. the 1-rd row, 5-th column in Fig. \ref{fig:supple_visualization} (b)), the corresponding attention weights in TAFA are very small (e.g. 0.032 in the 1-rd row, 5-rd column of Fig. \ref{fig:supple_visualization} (b)), making the irrelevant background features less influence the original ROI features.

%\vspace{-0.25cm}
\section{Conclusion}
%\vspace{-0.1cm}

%The proposed Temporal ROI Align is validated effectively on different baselines, different datasets and different vision tasks.

In this paper, a novel Temporal ROI Align is proposed to offer an alternative for the conventional ROI Align in video object detection. Compared with ROI Align which only extracts features of proposals from single-frame, the proposed operator can extract features from the entire video for proposals, making the extracted features contain long-range temporal information and much easier to perform classification for region classifier. The proposed operator is flexible and also validated effectively on different baselines, different datasets and different vision tasks about videos. We hope the proposed operator be applied into other video-related tasks, such as multi object tracking, in the future. 
%The proposed operator is flexible and can be easily integrated into other video detectors and other video tasks of computer vision, e.g. video instance segmentation.

\section{Acknowledgements}
This work is supported by the National Natural Science Foundation of China (Grant No. U20B2047, 62002336) and Exploration Fund Project of University of Science and Technology of China (YD3480002001).
%\clearpage

\bibliography{egbib}

\begin{thebibliography}{37}
\providecommand{\natexlab}[1]{#1}
\providecommand{\url}[1]{\texttt{#1}}
\providecommand{\urlprefix}{URL }
\expandafter\ifx\csname urlstyle\endcsname\relax
  \providecommand{\doi}[1]{doi:\discretionary{}{}{}#1}\else
  \providecommand{\doi}{doi:\discretionary{}{}{}\begingroup
  \urlstyle{rm}\Url}\fi

\bibitem[{Bertasius, Torresani, and Shi(2018)}]{bertasius2018object}
Bertasius, G.; Torresani, L.; and Shi, J. 2018.
\newblock Object detection in video with spatiotemporal sampling networks.
\newblock In \emph{Proceedings of the European Conference on Computer Vision
  (ECCV)}, 331--346.

\bibitem[{Cai and Vasconcelos(2018)}]{Cai_2018_CVPR}
Cai, Z.; and Vasconcelos, N. 2018.
\newblock Cascade r-cnn: Delving into high quality object detection.
\newblock In \emph{Proceedings of the IEEE Conference on Computer Vision and
  Pattern Recognition}, 6154--6162.

\bibitem[{Chen et~al.(2018)Chen, Wang, Yang, Zhang, Xiong, Change~Loy, and
  Lin}]{chen2018optimizing}
Chen, K.; Wang, J.; Yang, S.; Zhang, X.; Xiong, Y.; Change~Loy, C.; and Lin, D.
  2018.
\newblock Optimizing video object detection via a scale-time lattice.
\newblock In \emph{Proceedings of the IEEE conference on computer vision and
  pattern recognition}, 7814--7823.

\bibitem[{Chen et~al.(2020)Chen, Cao, Hu, and Wang}]{Chen_2020_CVPR}
Chen, Y.; Cao, Y.; Hu, H.; and Wang, L. 2020.
\newblock Memory Enhanced Global-Local Aggregation for Video Object Detection.
\newblock In \emph{IEEE/CVF Conference on Computer Vision and Pattern
  Recognition (CVPR)}.

\bibitem[{Dai et~al.(2016)Dai, Li, He, and Sun}]{dai2016r}
Dai, J.; Li, Y.; He, K.; and Sun, J. 2016.
\newblock R-fcn: Object detection via region-based fully convolutional
  networks.
\newblock In \emph{Proceedings of the Advances in Neural Information Processing
  Systems}, 379--387.

\bibitem[{Dai et~al.(2017)Dai, Qi, Xiong, Li, Zhang, Hu, and
  Wei}]{Dai_2017_ICCV}
Dai, J.; Qi, H.; Xiong, Y.; Li, Y.; Zhang, G.; Hu, H.; and Wei, Y. 2017.
\newblock Deformable convolutional networks.
\newblock In \emph{Proceedings of the IEEE International Conference on Computer
  Vision}, 764--773.

\bibitem[{Damen et~al.(2018)Damen, Doughty, Maria~Farinella, Fidler, Furnari,
  Kazakos, Moltisanti, Munro, Perrett, Price et~al.}]{damen2018scaling}
Damen, D.; Doughty, H.; Maria~Farinella, G.; Fidler, S.; Furnari, A.; Kazakos,
  E.; Moltisanti, D.; Munro, J.; Perrett, T.; Price, W.; et~al. 2018.
\newblock Scaling egocentric vision: The epic-kitchens dataset.
\newblock In \emph{Proceedings of the European Conference on Computer Vision
  (ECCV)}, 720--736.

\bibitem[{Deng et~al.(2019{\natexlab{a}})Deng, Hua, Song, Zhang, Xue, Ma,
  Robertson, and Guan}]{deng2019object}
Deng, H.; Hua, Y.; Song, T.; Zhang, Z.; Xue, Z.; Ma, R.; Robertson, N.; and
  Guan, H. 2019{\natexlab{a}}.
\newblock Object Guided External Memory Network for Video Object Detection.
\newblock In \emph{Proceedings of the IEEE International Conference on Computer
  Vision}, 6678--6687.

\bibitem[{Deng et~al.(2019{\natexlab{b}})Deng, Pan, Yao, Zhou, Li, and
  Mei}]{deng2019relation}
Deng, J.; Pan, Y.; Yao, T.; Zhou, W.; Li, H.; and Mei, T. 2019{\natexlab{b}}.
\newblock Relation Distillation Networks for Video Object Detection.
\newblock In \emph{Proceedings of the IEEE International Conference on Computer
  Vision}, 7023--7032.

\bibitem[{Dosovitskiy et~al.(2015)Dosovitskiy, Fischer, Ilg, Hausser, Hazirbas,
  Golkov, Van Der~Smagt, Cremers, and Brox}]{dosovitskiy2015flownet}
Dosovitskiy, A.; Fischer, P.; Ilg, E.; Hausser, P.; Hazirbas, C.; Golkov, V.;
  Van Der~Smagt, P.; Cremers, D.; and Brox, T. 2015.
\newblock Flownet: Learning optical flow with convolutional networks.
\newblock In \emph{Proceedings of the IEEE international conference on computer
  vision}, 2758--2766.

\bibitem[{Duan et~al.(2019)Duan, Bai, Xie, Qi, Huang, and
  Tian}]{duan2019centernet}
Duan, K.; Bai, S.; Xie, L.; Qi, H.; Huang, Q.; and Tian, Q. 2019.
\newblock Centernet: Keypoint triplets for object detection.
\newblock In \emph{Proceedings of the IEEE International Conference on Computer
  Vision}, 6569--6578.

\bibitem[{Feichtenhofer et~al.(2017)Feichtenhofer, Pinz, Zisserman, and
  Andrew}]{feichtenhofer2017detect}
Feichtenhofer, C.; Pinz, A.; Zisserman; and Andrew. 2017.
\newblock Detect to track and track to detect.
\newblock In \emph{Proceedings of the IEEE International Conference on Computer
  Vision}, 3038--3046.

\bibitem[{Girshick(2015)}]{girshick2015fast}
Girshick, R. 2015.
\newblock Fast R-CNN.
\newblock In \emph{Proceedings of the IEEE International Conference on Computer
  Vision}, 1440--1448.

\bibitem[{Girshick et~al.(2014)Girshick, Donahue, Darrell, and
  Malik}]{girshick2014rich}
Girshick, R.; Donahue, J.; Darrell, T.; and Malik, J. 2014.
\newblock Rich feature hierarchies for accurate object detection and semantic
  segmentation.
\newblock In \emph{Proceedings of the IEEE Conference on Computer Vision and
  Pattern Recognition}, 580--587.

\bibitem[{Gu et~al.(2018)Gu, Hu, Wang, Wei, and Dai}]{gu2018learning}
Gu, J.; Hu, H.; Wang, L.; Wei, Y.; and Dai, J. 2018.
\newblock Learning region features for object detection.
\newblock In \emph{Proceedings of the European Conference on Computer Vision
  (ECCV)}, 381--395.

\bibitem[{Guo et~al.(2019)Guo, Fan, Gu, Zhang, Xiang, Prinet, and
  Pan}]{guo2019progressive}
Guo, C.; Fan, B.; Gu, J.; Zhang, Q.; Xiang, S.; Prinet, V.; and Pan, C. 2019.
\newblock Progressive Sparse Local Attention for Video Object Detection.
\newblock In \emph{Proceedings of the IEEE International Conference on Computer
  Vision}, 3909--3918.

\bibitem[{Han et~al.(2016)Han, Khorrami, Paine, Ramachandran, Babaeizadeh, Shi,
  Li, Yan, and Huang}]{han2016seq}
Han, W.; Khorrami, P.; Paine, T.~L.; Ramachandran, P.; Babaeizadeh, M.; Shi,
  H.; Li, J.; Yan, S.; and Huang, T.~S. 2016.
\newblock Seq-nms for video object detection.
\newblock \emph{arXiv preprint arXiv:1602.08465} .

\bibitem[{He et~al.(2017)He, Gkioxari, Doll{\'a}r, and Girshick}]{He_2017_ICCV}
He, K.; Gkioxari, G.; Doll{\'a}r, P.; and Girshick, R. 2017.
\newblock Mask r-cnn.
\newblock In \emph{Proceedings of the IEEE International Conference on Computer
  Vision}, 2961--2969.

\bibitem[{He et~al.(2016)He, Zhang, Ren, and Sun}]{he2016deep}
He, K.; Zhang, X.; Ren, S.; and Sun, J. 2016.
\newblock Deep residual learning for image recognition.
\newblock In \emph{Proceedings of the IEEE Conference on Computer Vision and
  Pattern Recognition}, 770--778.

\bibitem[{Hu et~al.(2018)Hu, Gu, Zhang, Dai, and Wei}]{hu2018relation}
Hu, H.; Gu, J.; Zhang, Z.; Dai, J.; and Wei, Y. 2018.
\newblock Relation networks for object detection.
\newblock In \emph{Proceedings of the IEEE Conference on Computer Vision and
  Pattern Recognition}, 3588--3597.

\bibitem[{Lin et~al.(2017)Lin, Goyal, Girshick, He, and
  Doll{\'a}r}]{lin2017focal}
Lin, T.-Y.; Goyal, P.; Girshick, R.; He, K.; and Doll{\'a}r, P. 2017.
\newblock Focal loss for dense object detection.
\newblock In \emph{Proceedings of the IEEE international conference on computer
  vision}, 2980--2988.

\bibitem[{Liu and Zhu(2018)}]{liu2018mobile}
Liu, M.; and Zhu, M. 2018.
\newblock Mobile video object detection with temporally-aware feature maps.
\newblock In \emph{Proceedings of the IEEE Conference on Computer Vision and
  Pattern Recognition}, 5686--5695.

\bibitem[{Ren et~al.(2015)Ren, He, Girshick, and Sun}]{ren2015faster}
Ren, S.; He, K.; Girshick, R.; and Sun, J. 2015.
\newblock Faster R-CNN: Towards real-time object detection with region proposal
  networks.
\newblock In \emph{Proceedings of the Advances in Neural Information Processing
  Systems}, 91--99.

\bibitem[{Russakovsky et~al.(2015)Russakovsky, Deng, Su, Krause, Satheesh, Ma,
  Huang, Karpathy, Khosla, Bernstein et~al.}]{russakovsky2015imagenet}
Russakovsky, O.; Deng, J.; Su, H.; Krause, J.; Satheesh, S.; Ma, S.; Huang, Z.;
  Karpathy, A.; Khosla, A.; Bernstein, M.; et~al. 2015.
\newblock Imagenet large scale visual recognition challenge.
\newblock \emph{International journal of computer vision} 115(3): 211--252.

\bibitem[{Shvets et~al.(2019)Shvets, Liu, Berg, and C}]{shvets2019leveraging}
Shvets, M.; Liu, W.; Berg; and C, A. 2019.
\newblock Leveraging Long-Range Temporal Relationships Between Proposals for
  Video Object Detection.
\newblock In \emph{Proceedings of the IEEE International Conference on Computer
  Vision}, 9756--9764.

\bibitem[{Tian et~al.(2019)Tian, Shen, Chen, and He}]{tian2019fcos}
Tian, Z.; Shen, C.; Chen, H.; and He, T. 2019.
\newblock Fcos: Fully convolutional one-stage object detection.
\newblock In \emph{Proceedings of the IEEE International Conference on Computer
  Vision}, 9627--9636.

\bibitem[{Vaswani et~al.(2017)Vaswani, Shazeer, Parmar, Uszkoreit, Jones,
  Gomez, Kaiser, and Polosukhin}]{vaswani2017attention}
Vaswani, A.; Shazeer, N.; Parmar, N.; Uszkoreit, J.; Jones, L.; Gomez, A.~N.;
  Kaiser, {\L}.; and Polosukhin, I. 2017.
\newblock Attention is all you need.
\newblock In \emph{Advances in neural information processing systems},
  5998--6008.

\bibitem[{Wang et~al.(2018{\natexlab{a}})Wang, Zhou, Yan, and
  Deng}]{wang2018fully}
Wang, S.; Zhou, Y.; Yan, J.; and Deng, Z. 2018{\natexlab{a}}.
\newblock Fully motion-aware network for video object detection.
\newblock In \emph{Proceedings of the European Conference on Computer Vision
  (ECCV)}, 542--557.

\bibitem[{Wang et~al.(2018{\natexlab{b}})Wang, Girshick, Gupta, and
  He}]{wang2018non}
Wang, X.; Girshick, R.; Gupta, A.; and He, K. 2018{\natexlab{b}}.
\newblock Non-local neural networks.
\newblock In \emph{Proceedings of the IEEE conference on computer vision and
  pattern recognition}, 7794--7803.

\bibitem[{Wu et~al.(2019)Wu, Chen, Wang, and Zhang}]{wu2019sequence}
Wu, H.; Chen, Y.; Wang, N.; and Zhang, Z. 2019.
\newblock Sequence Level Semantics Aggregation for Video Object Detection.
\newblock In \emph{Proceedings of the IEEE International Conference on Computer
  Vision}, 9217--9225.

\bibitem[{Xiao and Jae~Lee(2018)}]{xiao2018video}
Xiao, F.; and Jae~Lee, Y. 2018.
\newblock Video object detection with an aligned spatial-temporal memory.
\newblock In \emph{Proceedings of the European Conference on Computer Vision
  (ECCV)}, 485--501.

\bibitem[{Xie et~al.(2017)Xie, Girshick, Doll{\'a}r, Tu, and
  He}]{xie2017aggregated}
Xie, S.; Girshick, R.; Doll{\'a}r, P.; Tu, Z.; and He, K. 2017.
\newblock Aggregated residual transformations for deep neural networks.
\newblock In \emph{Proceedings of the IEEE conference on computer vision and
  pattern recognition}, 1492--1500.

\bibitem[{Yang, Fan, and Xu(2019)}]{yang2019video}
Yang, L.; Fan, Y.; and Xu, N. 2019.
\newblock Video instance segmentation.
\newblock In \emph{Proceedings of the IEEE International Conference on Computer
  Vision}, 5188--5197.

\bibitem[{Zhu et~al.(2018)Zhu, Dai, Yuan, and Wei}]{zhu2018towards}
Zhu, X.; Dai, J.; Yuan, L.; and Wei, Y. 2018.
\newblock Towards high performance video object detection.
\newblock In \emph{Proceedings of the IEEE Conference on Computer Vision and
  Pattern Recognition}, 7210--7218.

\bibitem[{Zhu et~al.(2019)Zhu, Hu, Lin, and Dai}]{zhu2019deformable}
Zhu, X.; Hu, H.; Lin, S.; and Dai, J. 2019.
\newblock Deformable convnets v2: More deformable, better results.
\newblock In \emph{Proceedings of the IEEE Conference on Computer Vision and
  Pattern Recognition}, 9308--9316.

\bibitem[{Zhu et~al.(2017{\natexlab{a}})Zhu, Wang, Dai, Yuan, and
  Wei}]{zhu2017flow}
Zhu, X.; Wang, Y.; Dai, J.; Yuan, L.; and Wei, Y. 2017{\natexlab{a}}.
\newblock Flow-guided feature aggregation for video object detection.
\newblock In \emph{Proceedings of the IEEE International Conference on Computer
  Vision}, 408--417.

\bibitem[{Zhu et~al.(2017{\natexlab{b}})Zhu, Xiong, Dai, Yuan, and
  Wei}]{zhu2017deep}
Zhu, X.; Xiong, Y.; Dai, J.; Yuan, L.; and Wei, Y. 2017{\natexlab{b}}.
\newblock Deep feature flow for video recognition.
\newblock In \emph{Proceedings of the IEEE Conference on Computer Vision and
  Pattern Recognition}, 2349--2358.

\end{thebibliography}

\end{document}